\PassOptionsToPackage{table,dvipsnames}{xcolor}

\documentclass{maskwam}

\usepackage{amsmath}
\usepackage{amssymb}
\usepackage{amsfonts}
\usepackage{mathtools}
\usepackage{enumitem}
\usepackage{wrapfig}
\usepackage{adjustbox}
\usepackage{tabularx}
\usepackage{makecell}
\usepackage{siunitx}
\usepackage{float}
\usepackage{diagbox}
\usepackage{dsfont}

\usepackage{array}
\usepackage{colortbl}
\usepackage{amsthm}
\usepackage{pifont}
\usepackage{marvosym}

\newtheorem{assumption}{Assumption}

\newcommand{\cmark}{\ding{51}}
\newcommand{\xmark}{\ding{55}}

\newcommand{\question}{%
    \stepcounter{question}%
    \noindent\textbf{Q\thequestion:~\ignorespaces}%
}

\newcounter{question}
\definecolor{baselinecolor}{gray}{.9}

\newcolumntype{x}[1]{>{\centering\arraybackslash}p{#1pt}}

\definecolor{drp-blue}{HTML}{1f77b4}
\definecolor{pretty-blue}{RGB}{0, 113, 188}
\definecolor{kaiming-green}{RGB}{57,181,74}
\definecolor{mypurple}{RGB}{55,0,168}
\definecolor{icmlblue}{rgb}{0,0.08,0.45}
\definecolor{mygreen}{HTML}{4FC978}
\definecolor{linecolor1}{RGB}{246, 248, 239}
\definecolor{linecolor2}{RGB}{230, 234, 217}
\definecolor{linecolor3}{RGB}{211, 222, 190}
\definecolor{reconcolor}{HTML}{412F8A}
\definecolor{runpei-orange}{HTML}{F35F27}
\definecolor{runpei_blue}{HTML}{14294B}
\definecolor{datacolor}{HTML}{0009BF}
\definecolor{vitcolor}{HTML}{fc8e62}
\definecolor{cvprblue}{rgb}{0.21,0.49,0.74}
\definecolor{myblue}{rgb}{.39,.58,.93}

\newcommand{\method}{DeformGen}

\title{DeformGen: Dynamics-Based Topology Augmentation for Deformable Manipulation Policy Learning}

\renewcommand{\authorlist}{%
  {\sffamily \sbseries
  Zili Lin       $^{12}$,
  Wenyao Zhang   $^{12\dagger}$,
  Yuyang Zhang   $^{12}$,
  Zekun Qi       $^{3}$,
  Junyan Lin     $^{24}$,
  Hanxin Zhu     $^{56}$,\\
  Jiaolong Yang,
  Zhibo Chen     $^{56}$,
  Yao Mu         $^{1}$,
  Xiaokang Yang  $^{1}$,
  Xin Jin        $^{26}$,
  Wenjun Zeng    $^{2\text{\Letter}}$
  }%
}

\affiliation[1]{Shanghai Jiao Tong University}
\affiliation[2]{Eastern Institute of Technology, Ningbo}
\affiliation[3]{Tsinghua University}
\affiliation[4]{The Hong Kong Polytechnic University}
\affiliation[5]{University of Science and Technology of China}
\affiliation[6]{Zhongguancun Academy}

\contribution[\dagger]{Project Lead}
\contribution[\text{\Letter}]{Corresponding author}

\abstract{
Demonstration augmentation is proposed for cost-efficient data acquisition, but existing methods are fundamentally limited in deformable manipulation due to two challenges: (1)~the \emph{state space} is high-dimensional with physics-induced constraints, making valid configurations impossible to reach via low-dimensional pose perturbations; and (2)~\emph{trajectory transfer} is non-equivariant, as material points no longer move rigidly together under deformation.
We present \textbf{DeformGen}, a dynamics-based augmentation framework that achieves \emph{topological diversity} for deformable objects. For the state challenge, DeformGen expands the valid state distribution by applying localized physical disturbances and forward-simulating the dynamics to obtain topology-coherent, physically plausible deformable states.
For the trajectory challenge, DeformGen transfers source manipulation trajectories via deformation-field warping, which lifts per-particle displacements into a continuous spatial function to adapt the end-effector trajectory consistently with the deformed geometry.
In this way, our method jointly augments the state distribution and its associated manipulation behavior.
Experiments on high-fidelity deformable manipulation benchmarks show that DeformGen generally improves policy learning compared with training on the original demonstrations alone and with rigid-style augmentation baselines.
}

\date{\sffamily\today}
\metadata[Project Page]{\url{https://zili2002.github.io/DeformGen}}
\metadata[Github]{\url{https://github.com/Zili2002/DeformGen}}

\begin{document}
\maketitle
\vspace{-5mm}

\begin{figure}[h!]
    \centering
    \includegraphics[width=0.99\linewidth]{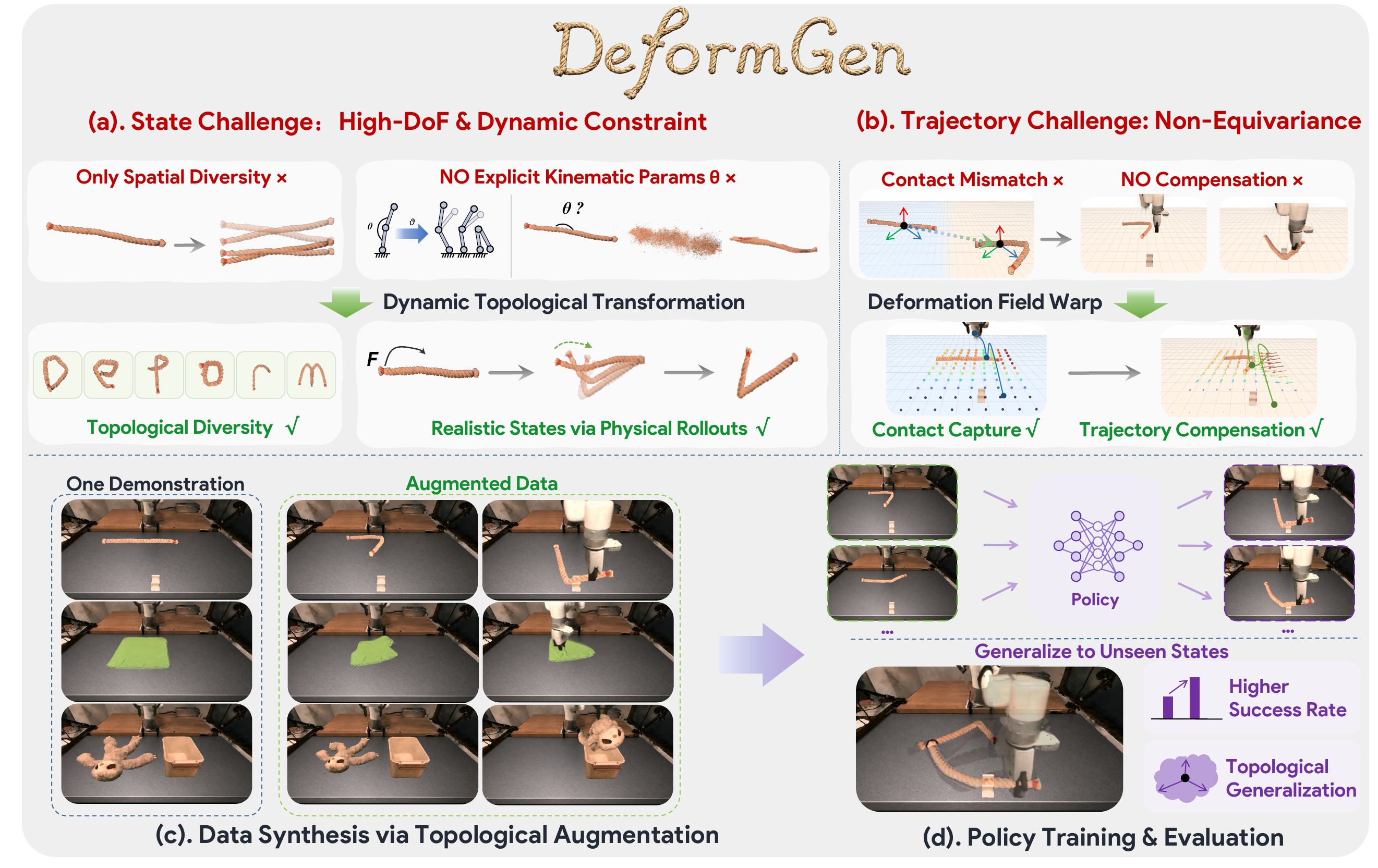}
    \vspace{-3mm}
    \caption{
\textit{Top:} We identify two core challenges -- the state-space challenge and the trajectory-transfer challenge -- that prevent rigid-style augmentation from extending to deformable manipulation. \textbf{DeformGen} addresses them through dynamics-based topology transformation and deformation-field warping.
\textit{Bottom:} Starting from a single demonstration, DeformGen synthesizes diverse demonstrations across deformable states, leading to improved policy generalization to unseen states.
}
    \label{fig:teaser}
\end{figure}
\vspace{-1mm}

\section{Introduction}

Imitation learning and visuomotor policy learning have shown remarkable success in robot manipulation, enabling policies to conduct various tasks across diverse environments~\cite{24pi0,25pi0.5,kim2024openvla,bjorck2025gr00t,chen2025internvla,zhang2025dreamvla,zhang2026disentangled,sun2026vla,liang2025discrete}. 
However, this progress has been driven in large part by access to large-scale, diverse demonstration data, whose collection remains expensive, time-consuming, and difficult to scale. 
To mitigate this bottleneck, a data augmentation paradigm has emerged: rather than collecting or synthesizing more demonstrations~\cite{mu2025robotwin}, these methods typically augment a single human demonstration to many training trajectories~\cite{mandlekar2023mimicgen,xue2025demogen,yang2025novel,xu2025egodemogen}. 
This idea is based on the simple but powerful \emph{equivariant assumption}: 
rigid bodies satisfy the equivariant constraint that the Euclidean distance between any two material points is invariant under motion or contact forces.
With the same rigid transformation applied to the end-effector, the relative pose between it and the object is preserved, leading to a valid trajectory.

This augmentation paradigm, however, is fundamentally mismatched to deformable object manipulation~\cite{moghani2026softmimicgen,zhou2026sim1}. 
As illustrated in Fig.~\ref{fig:teaser}, we identify two core challenges that break the rigid-object recipe.
\textbf{State-Space Challenge(Fig.~\ref{fig:teaser}(a)):} (i) \textit{High degrees of freedom.} For rigid bodies, a 6-DoF pose provides a sufficient state abstraction. 
Deformable objects, in contrast, exhibit rich, high-dimensional shape and topology variations~\cite{sanchez2018robotic,yin2021modeling}. 
As a result, rigid transformations alone cannot meaningfully expand the valid state distribution required for deformable manipulation.
(ii) \textit{Dynamic constraint.}
For rigid and articulated objects, valid perturbations can be constructed directly in pose or joint space via kinematic constraints.
For deformable objects, however, internal constraints are dynamic, which means the deformation depends on the interaction between internal particles; therefore, naive geometric perturbations typically yield implausible shapes and discontinuous structural changes.

\textbf{Trajectory-Transfer Challenge(Fig.~\ref{fig:teaser}(b)):} 
The \emph{equivariant assumption} no longer holds for deformable objects: material points are not equivariant~\cite{moghani2026softmimicgen}. 
Consequently, directly applying a global rigid isometry to augment the trajectories for topological variants of deformable objects introduces two problems: (i) the grasp pose becomes misaligned with the object's local geometry, so the end-effector can no longer grip the object correctly; and (ii) a rigid-style trajectory transfer can only translate and rotate the trajectory as a whole, and cannot capture or compensate for the object's local deformation. 
These challenges suggest that effective augmentation for deformable manipulation must jointly address two problems: synthesizing physically valid deformable states and transferring demonstrations in a deformation-aware manner. 

To this end, we propose \textbf{DeformGen}, a dynamics-based topology data augmentation framework for deformable manipulation. 

Unlike prior rigid augmentation methods that are confined to SE(3) perturbations and thus only produce spatial diversity, DeformGen achieves effective \emph{topological} diversity for deformable objects by jointly synthesizing physically valid deformed states and transferring demonstrations in a deformation-aware manner.
Specifically, for the state-space challenge, the key insight is that physically plausible states form a constrained manifold within the high-dimensional particle state space, and naive geometric perturbations almost always fall off this manifold.
Therefore, we propose \emph{Dynamic Topological Transformation} to augment the state distribution by applying randomized, spatially localized forces to the object and forward-simulating the resulting dynamics to prevent leaving the valid manifold.

These augmented assets can be used both to enrich the support of training demonstrations and to broaden policy evaluation beyond the narrow state distribution covered by the original data.

In response to the trajectory-transfer challenge, \method{} transfers source demonstrations to each augmented state via \emph{Deformation-Field Warping}.
We compute per-particle displacements between the source and target object states and lift them, through $K$-nearest-neighbor inverse-distance interpolation, into a continuous deformation field $D(\mathbf{x})$ over the workspace.
Applying $D$ to the demonstration trajectory simultaneously re-orients the gripper pose according to local geometric changes near the grasp region and compensates the global trajectory to remain aligned with the deformed object as a whole.
A single demonstration then can be reused across an entire family of deformable states without breaking contact or violating the object's physical structure(Fig.~\ref{fig:teaser}(c)). 

As shown in Fig.~\ref{fig:teaser}(d), we train policies on the resulting augmented dataset to improve generalization and robustness. 
Experiments on high-fidelity deformable manipulation benchmark~\cite{zhang2025real2sim} show that DeformGen 
 generally improves policy learning: compared with training on the original demonstrations alone, or with rigid-style augmentation baselines, policies trained with DeformGen achieve 
 higher success rates in most settings. These results suggest that effective augmentation for deformable manipulation requires dynamics-consistent state synthesis coupled with deformation-aware trajectory transfer, rather than rigid pose perturbation alone.
The contributions of this work are three-fold:
\begin{itemize}[leftmargin=1.5em,topsep=1.5pt,itemsep=1.5pt]
    \item Formulation of demonstration augmentation for deformable manipulation that identifies physically valid deformable state synthesis as the key missing ingredient beyond rigid-style augmentation.
    \item A dynamics-consistent pipeline that generates topology-coherent deformable assets through localized perturbation, physics rollout, and stabilization, and synthesizes corresponding manipulation trajectories via deformation-field warping;
    \item Extensive empirical evidence that the resulting synthetic demonstrations significantly improve policy learning, providing gains over both no augmentation and rigid-style augmentation baselines.
\end{itemize}
\vspace{-3mm}

\section{Related Works}
\vspace{-2mm}

\subsection{Data augmentation for robot manipulation}
\vspace{-1mm}
Unlike pipelines that generate demonstrations from scratch using planners~\cite{High-Fidelity}, generative models, or learned agents~\cite{james2019rlbench,wang2024robogen,kanehira2025rldriven}, data augmentation expands an existing dataset through task-agnostic transformations~\cite{mandlekar2023mimicgen,xue2025demogen}.
Beyond appearance-only \emph{visual} perturbations~\cite{chen2025semantically,gigaai2025gigabrain0}, \emph{behavioral} augmentation modifies object configurations and re-solves for task-successful trajectories, typically via physics-based planning~\cite{mandlekar2023mimicgen,jiang2025dexmimicgen,xue2025demogen,yang2025novel,xu2025egodemogen}, image/video generation models~\cite{jang2025dreamgen,li2026manipdreamer3d,ji2025oxe,wang2026robovip} and Real2sim2real~\cite{pan2025one,r2r2r}.
Within the physics-based line, DemoGen~\cite{xue2025demogen} edits 3D point clouds directly and its extension R2E2R~\cite{zhao2025real2edit2real} renders consistent videos via a depth-conditioned generator.
Simulation-based variants~\cite{mandlekar2023mimicgen,garrett2024skillmimicgen,jiang2025dexmimicgen,r2r2r} augment via geometric transformations in a digital twin, extending to clutter, bimanual embodiments, and photorealistic 3DGS~\cite{kerbl20233d} rendering.
However, they rely on rigid SE(3) transformations that break down on deformables. 
Unlike DeformGen, SoftMimicGen~\cite{moghani2026softmimicgen} has tried to mitigate the trajectory-transfer challenge, but it still faces the limitation of state-space.

\subsection{Deformable object manipulation}
\vspace{-1mm}
Early works use model-based planning built on physical simulators or learned dynamics models, using mass--spring~\cite{springmass}, FEM~\cite{FEM}, or particle-based representations~\cite{MPM,PBD,orozco2025learning} to predict deformation and plan actions~\cite{lin2022diffskill,shi2023robocook,chen2023predicting,huang2023defgraspsim,han2026robotic,mckennaa2026perspective}.
A second line of work removes the need for explicit dynamics by directly learning visuomotor policies~\cite{chi2024diffusion,ACT23} from demonstrations or interaction, covering tasks such as knot tying~\cite{peng2024tiebot}, cable insertion~\cite{wu2025robotic}, cloth folding~\cite{sima2026kai0,seita2020deep,weng2022fabricflownet,wang2025dexgarmentlab,ha2022flingbot}, and dough shaping~\cite{shi2023robocook}.
More recently, large-scale vision--language--action models have been extended to deformable settings, showing promising generalization but requiring more data than their rigid-object counterparts~\cite{bjorck2025gr00t,chen2025internvla,zhang2026disentangled}.
However, this progress has been driven in large part by access to large-scale, diverse demonstration data, whose collection remains expensive, time-consuming, and difficult to scale.

\section{Method}

In this section, we propose \method{}, which aims to synthesize a large volume of valid manipulation data for the same task but with varying object initial states, starting from sparse demonstration data. 
First, we present a novel object initial state augmentation approach in Sec.~\ref{subsec:state_augmentation}.
Based on the above deformable object representation, we propose a manipulation trajectory augmentation method in Sec.~\ref{subsec:tracjectory_augmentation}.
Furthermore, we describe the policy training framework employed to verify the effectiveness and efficiency of the augmented data in Sec.~\ref{subsec:policy_learning}.
For soft object modeling and simulation, we leverage PhysTwin~\cite{jiang2025phystwin} and Real2Sim-Eval~\cite{zhang2025real2sim} due to their high fidelity in both visual rendering and physical dynamics, with detailed descriptions provided in Appendix~\ref{app:simulation_details}.

\subsection{State Augmentation}
\label{subsec:state_augmentation}
\begin{figure}[tbp]
    \centering
    \includegraphics[width=0.9\textwidth]{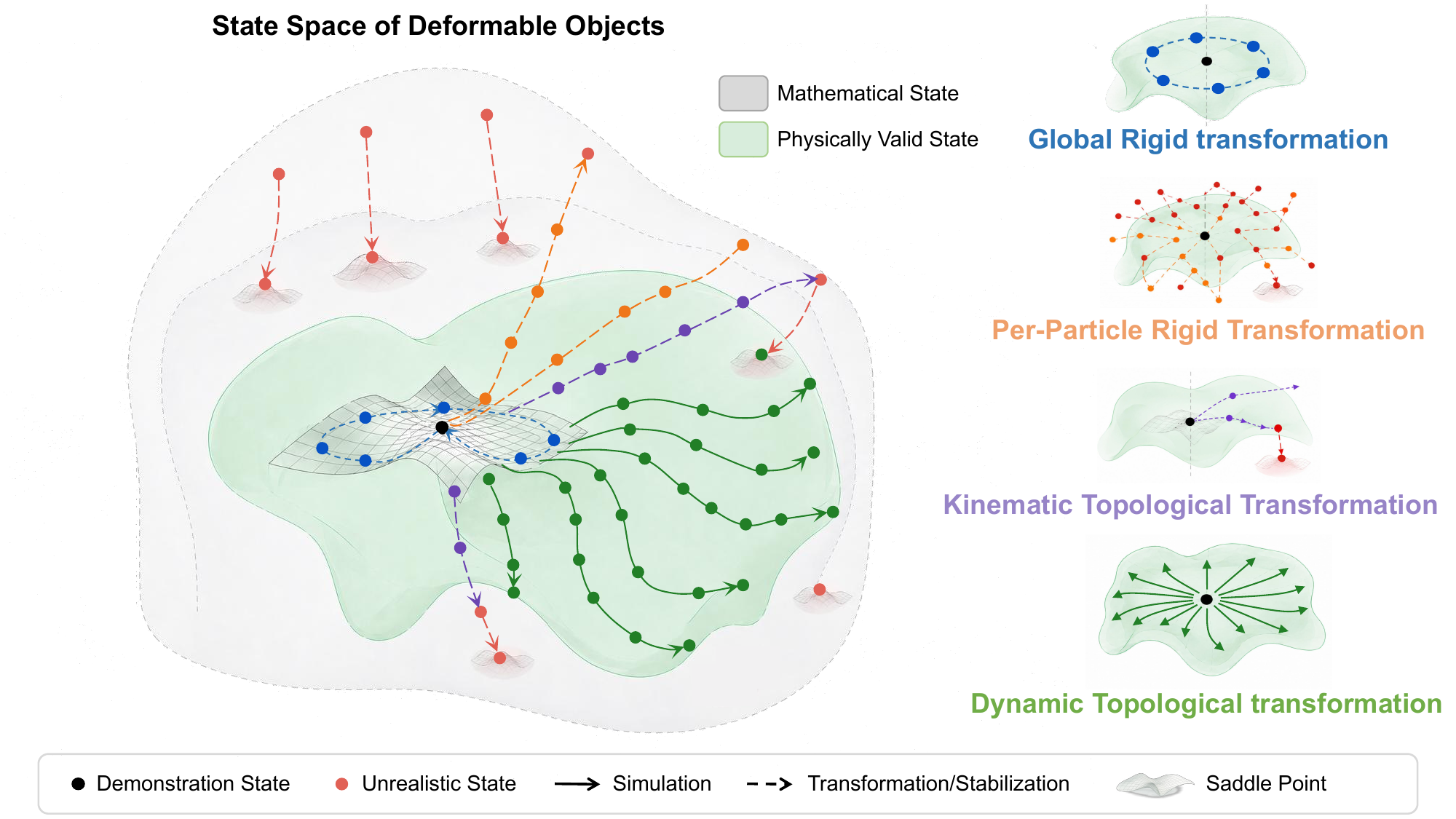}
    \caption{\textbf{Augmentation strategies in deformable state space.}
    Each strategy is visualized in the configuration space $\mathcal{S}$ with
    the physically plausible subspace $\mathcal{S}_{\mathrm{real}}$ shaded.
    Dynamics-based augmentation is designed to keep all generated states
    within~$\mathcal{S}_{\mathrm{real}}$ while achieving broader coverage than alternatives.}
    ~\vspace{-8mm}
    \label{fig:state_space}
\end{figure}

The objective of this step is to generate diverse object configurations for the
same task, serving both subsequent trajectory synthesis and policy evaluation.
Fundamentally, synthesizing object states amounts to sampling the object's
configuration space.
A practical method should produce states that are physically plausible under the simulator's dynamics model and that are sufficiently diverse to improve downstream policy learning.

\paragraph{State space.}
Following standard practice in deformable object simulation~\cite{MPM,PBD,corl2020softgym,hu2019taichi,springmass,macklin2022warp}, we consider a deformable object discretized into $N$ particles with configuration space $\mathcal{S} = \mathbb{R}^{3N}$, where each state $\mathbf{s} = (\mathbf{p}_1, \dots, \mathbf{p}_N) \in \mathcal{S}$ specifies all particle positions. The \emph{physically plausible subspace} $\mathcal{S}_{\mathrm{real}} \subset \mathcal{S}$ contains all configurations consistent with real-world physical constraints. In general, $\mathcal{S}_{\mathrm{real}} \subsetneq \mathcal{S}$: most points in $\mathbb{R}^{3N}$ do not correspond to any physically realizable configuration.

\paragraph{Working assumption.}
Our approach relies on the premise that a well-calibrated physics simulator $\Phi_{\mathrm{sim}}(\mathbf{s}, \mathbf{f}, \Delta t)$ approximately preserves physical plausibility when evolving from a valid state, but cannot reliably restore it from an invalid one (Assumption~\ref{ass:forward-invariance} in Appendix~\ref{app:state_aug_details}). This asymmetry implies that any method which first perturbs the state out of $\mathcal{S}_{\mathrm{real}}$ and then relies on simulation to ``fix'' it has no reliable path back to plausibility.

\paragraph{Why existing strategies fall short.}
We identify three alternatives (details in Appendix~\ref{app:augmentation_analysis}):
(i)~\emph{Global rigid transformation} preserves plausibility but is confined to a 6-DoF subspace of~$\mathbf{s}_0$, unable to capture shape or topological variation---confirmed by near-zero non-rigid residuals in Fig.~\ref{fig:state_analysis}.
(ii)~\emph{Per-particle perturbation} faces a coverage--plausibility trade-off: large noise breaks connectivity; small noise yields only local wrinkles~\cite{tian2025interndata}.
(iii)~\emph{Kinematic deformation fields} preserve topology but ignore material constraints, producing coherent yet dynamically inadmissible states.
Both~(ii) and~(iii) leave~$\mathcal{S}_{\mathrm{real}}$ and rely on post-hoc repair that may fail.

\begin{figure}[!htbp]
    \centering
    \includegraphics[width=\textwidth]{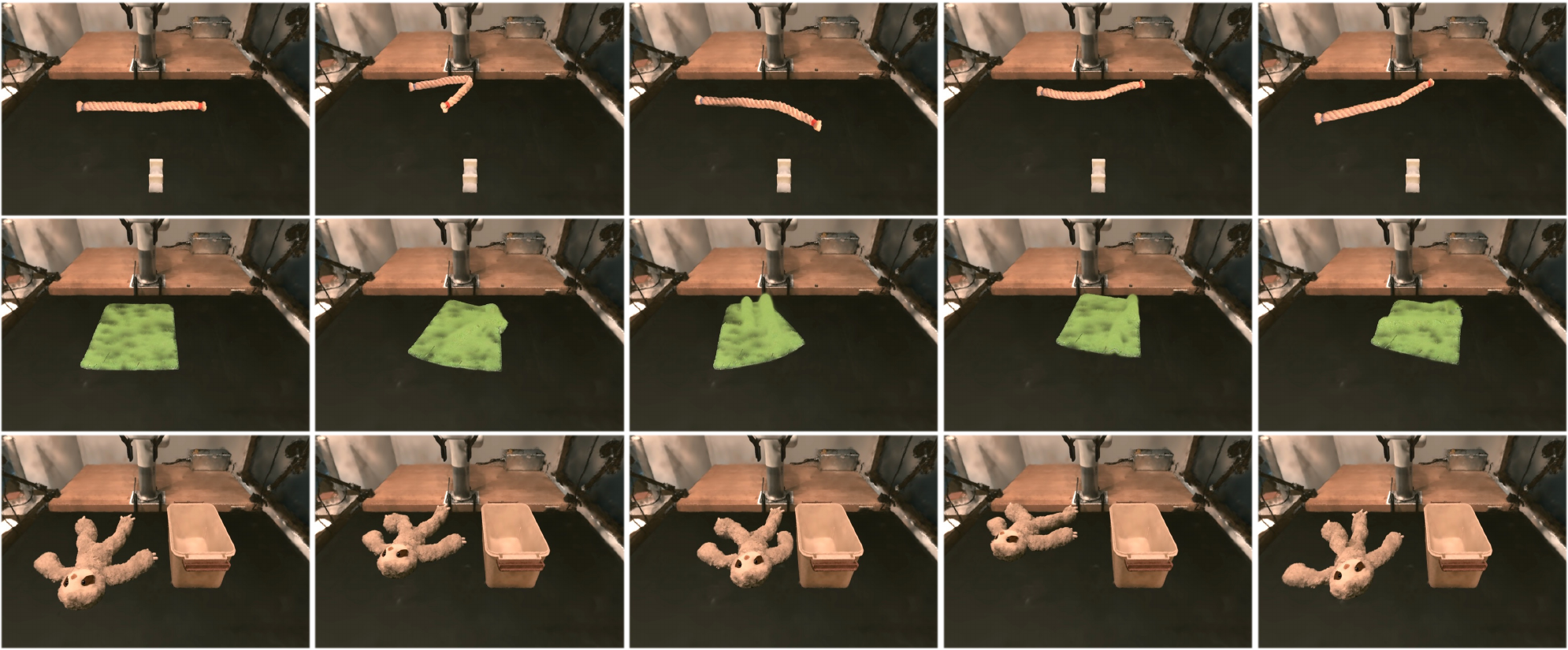}
    \vspace{-5mm}
    \caption{\textbf{Examples of augmented object states.} Each row shows one task. The leftmost column is the source demonstration state; subsequent columns show states generated by DeformGen via dynamics-based topological augmentation. All states are physically plausible and exhibit diverse topological variations.}
    \label{fig:state_aug_examples}
    \vspace{-6mm}
\end{figure}

\paragraph{Dynamics-based topological augmentation.}
We instead augment states by applying localized external forces and \emph{forward-simulating} the dynamics from a known valid state:
\begin{equation}
  \mathbf{s}_{\mathrm{aug}}
  = \Phi_{\mathrm{sim}}(\mathbf{s}_0,\, \mathbf{f},\, \Delta t),
  \quad \mathbf{s}_0 \in \mathcal{S}_{\mathrm{real}},
  \label{eq:dynamics_aug}
\end{equation}
where $\mathbf{f}$ is a localized force field. Because the method evolves the state through the simulator's own dynamics, it never explicitly leaves~$\mathcal{S}_{\mathrm{real}}$, requiring no post-hoc repair from invalid states. Since localized forces can induce diverse non-rigid deformations (bending, twisting, folding, draping), the reachable set is not restricted to a low-dimensional submanifold. We do not claim full coverage of~$\mathcal{S}_{\mathrm{real}}$, but treat this as a \emph{practical sampling heuristic} that explores a substantially broader region than rigid transformations---verified empirically in Fig.~\ref{fig:state_analysis}. Table~\ref{tab:augmentation-comparison} summarizes the comparison.

\begin{table}[htbp]
  \centering
  \caption{Comparison of augmentation strategies for deformable objects.
  \emph{Coherence}: preserves topological coherence.
  $\subseteq \mathcal{S}_{\mathrm{real}}$: reachable states remain plausible under Assumption~\ref{ass:forward-invariance}.}
  \label{tab:augmentation-comparison}
  \resizebox{\textwidth}{!}{%
  \setlength{\tabcolsep}{15pt}
  \begin{tabular}{lcccc}
    \toprule
    \textbf{Strategy}
      & \textbf{Coherence}
      & \textbf{Reachable set}
      & $\boldsymbol{\subseteq \mathcal{S}_{\mathrm{real}}}$
      & \textbf{~$\boldsymbol{\mathcal{S}_{\mathrm{real}}}$-recoverable?} \\
    \midrule
    (i)\; Global rigid
      & \cmark
      & 6-DoF subspace of $\mathbf{s}_0$
      & \cmark
      & N/A \\
    (ii)\; Per-particle
      & \xmark
      & $\mathcal{S}$
      & \xmark
      & Unreliable \\
    (iii)\; Kinematic topological
      & \cmark
      & $\mathcal{S}$
      & \xmark
      & Unreliable \\
    \textbf{(iv)\; Dynamics (Ours)}
      & \textbf{\cmark}
      & $\boldsymbol{\mathcal{R}(\mathbf{s}_0) \subseteq \mathcal{S}_{\mathrm{real}}}$
      & \textbf{\cmark\textsuperscript{$\dagger$}}
      & \textbf{N/A} \\
    \bottomrule
    \multicolumn{5}{l}{\small \textsuperscript{$\dagger$}Under Assumption~\ref{ass:forward-invariance}; in practice, subject to simulator fidelity.}
  \end{tabular}
}
\end{table}

\subsection{Trajectory Augmentation}
\label{subsec:tracjectory_augmentation}
\begin{figure}[tbp]
    \vspace{-2mm}
    \centering
    \includegraphics[width=1.0\textwidth]{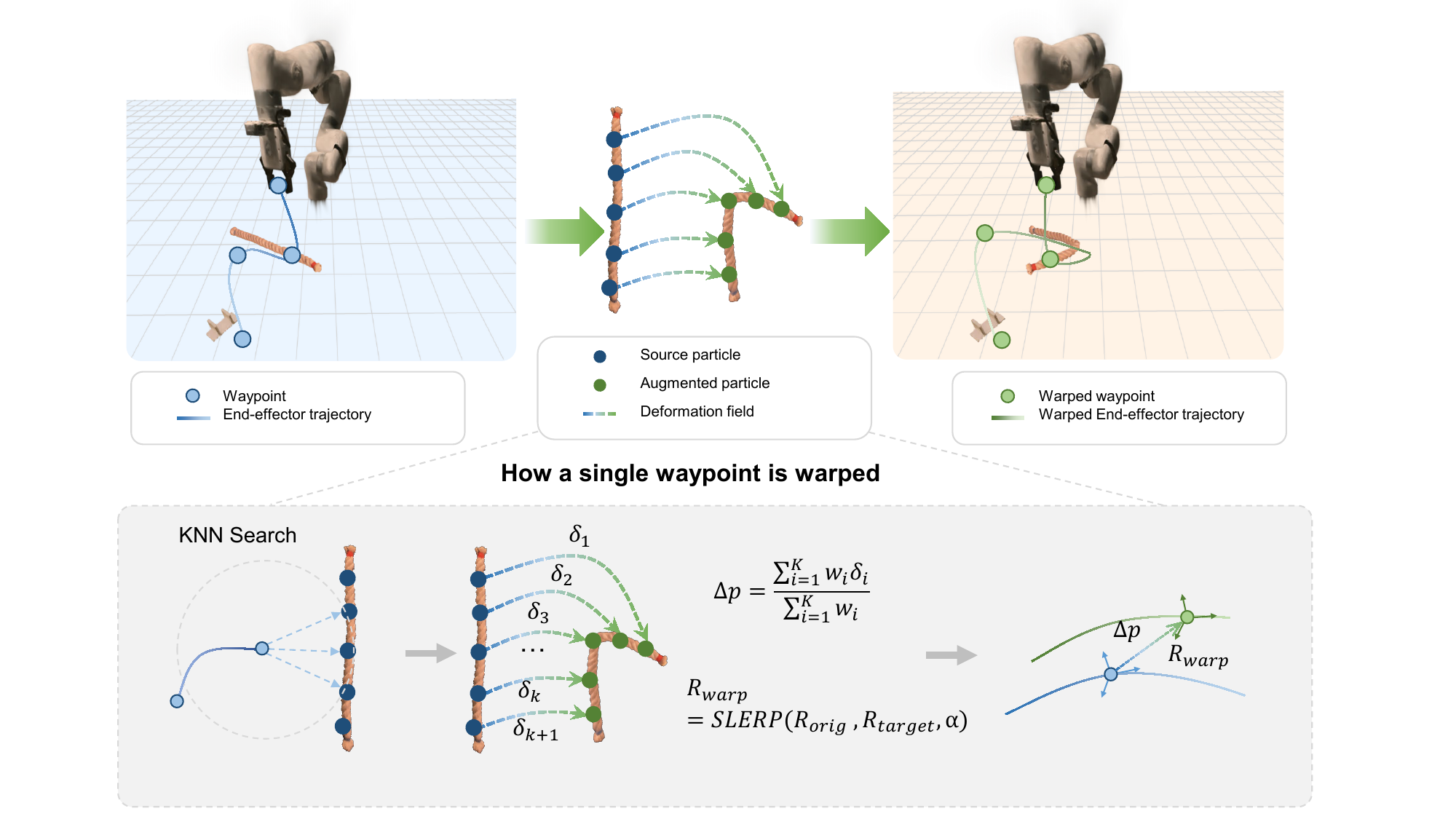}
    \vspace{-3mm}
    \caption{\textbf{Trajectory augmentation via deformation-field warping.} \textbf{Top:} The source trajectory (\textcolor{blue}{blue}, left) is warped through the deformation field (\textcolor{cyan}{dashed arrows}, center) to produce an augmented trajectory (\textcolor{green}{green}, right) consistent with the deformed object. \textbf{Bottom:} For each waypoint, $K$-nearest-neighbor particle displacements are aggregated via inverse-distance weighting to obtain the position offset $\Delta p$, and a local Jacobian is estimated to derive the orientation update $R_{\mathrm{warp}}$ via SLERP.}
    \label{fig:tracjectory_warp}
    \vspace{-3mm}
\end{figure}

The objective of this step is to synthesize valid manipulation trajectories for unseen object configurations. Our synthesized trajectories consist of three phases: \textit{approach}, \textit{grasp}, and \textit{manipulation}. The grasp poses and manipulation trajectories are synthesized using \textit{Deformation Field Warping}, while the approach trajectory is generated via interpolation from the robot's reset pose to the grasp pose.

Rigid trajectory transfer methods~\cite{xue2025demogen,r2r2r} assume uniform transformation across the object, neglecting distinct deformations across different parts of a deformable object. Inspired by~\cite{schulman2016learning}, we construct a deformation field from per-particle displacements, yielding a closed-form spatial mapping without iterative optimization.

\paragraph{Position warping.}
Let $p_{\mathrm{orig}}, p_{\mathrm{def}} \in \mathbb{R}^{N \times 3}$ be the source and deformed point clouds. The per-point displacement is $\delta_i = p_{\mathrm{def},i} - p_{\mathrm{orig},i}$.
For each end-effector position $x_t$ at timestep $t$, we retrieve its $k$ nearest neighbors from $p_{\mathrm{orig}}$ and interpolate via inverse distance weighting:
\begin{equation}
    w_{t,j} = \frac{1}{\left\|x_t - p_{\mathrm{orig},\,\mathrm{nn}_j(x_t)}\right\| + \varepsilon}, \ \
    \tilde{w}_{t,j} = \frac{w_{t,j}}{\sum_j w_{t,j}}, \ \
    d(x_t) = \sum_j \tilde{w}_{t,j}\,\delta_{\mathrm{nn}_j(x_t)},
\end{equation}
where $\varepsilon > 0$ ensures numerical stability. The warped position incorporates a time-dependent decay:
\begin{equation}
    x_t^{\mathrm{warp}} = x_t + \alpha_t \cdot d(x_t),
\end{equation}
where $\alpha_t = \mathrm{decay}(t)$ allows the trajectory to follow local deformations initially while gradually reverting to the original path.

\paragraph{Orientation adaptation.}
For the end-effector orientation, we construct local relative coordinates within the KNN neighborhood of $x_t$:
\begin{equation}
    \ell^{\mathrm{orig}}_{t,j} = p_{\mathrm{orig},\,\mathrm{nn}_j(x_t)} - x_t, \quad
    \ell^{\mathrm{def}}_{t,j} = \ell^{\mathrm{orig}}_{t,j} + \delta_{\mathrm{nn}_j(x_t)}.
\end{equation}
A local Jacobian matrix $J_t$ is estimated via least squares fitting to map the original local vectors to the deformed ones:
\begin{equation}
    J_t = \arg\min_J \sum_j \left\| \ell^{\mathrm{def}}_{t,j} - J\,\ell^{\mathrm{orig}}_{t,j} \right\|^2.
\end{equation}
Letting $X_{\mathrm{orig}}$ and $X_{\mathrm{def}}$ denote the matrices of stacked local vectors, the closed-form solution is:
\begin{equation}
    J_t = X_{\mathrm{def}} X_{\mathrm{orig}}^{\top} \left( X_{\mathrm{orig}} X_{\mathrm{orig}}^{\top} \right)^{+},
\end{equation}
where $(\cdot)^{+}$ denotes the Moore-Penrose pseudoinverse. The induced rotation $R_t'$ is obtained by projecting $J_t R_t$ onto the $SO(3)$ manifold via SVD. The final warped orientation is computed via:
\begin{equation}
    R_t^{\mathrm{warp}} = \mathrm{SLERP}\left(R_t,\; R_t',\; \alpha_t\right).
\end{equation}

In practice, the grasp pose correlates more strongly with nearby object points, so we use a small $K$ for warping the grasp pose. The manipulation phase depends on the overall object state, so we set $K$ to the total number of object points to capture global deformation. Given the tabletop scenario, we constrain rotations to the Z-axis perpendicular to the table surface. Details are in Appendix~\ref{app:trajectory_details}.

\begin{figure}[tbp]
    \centering
    \includegraphics[width=0.95\textwidth]{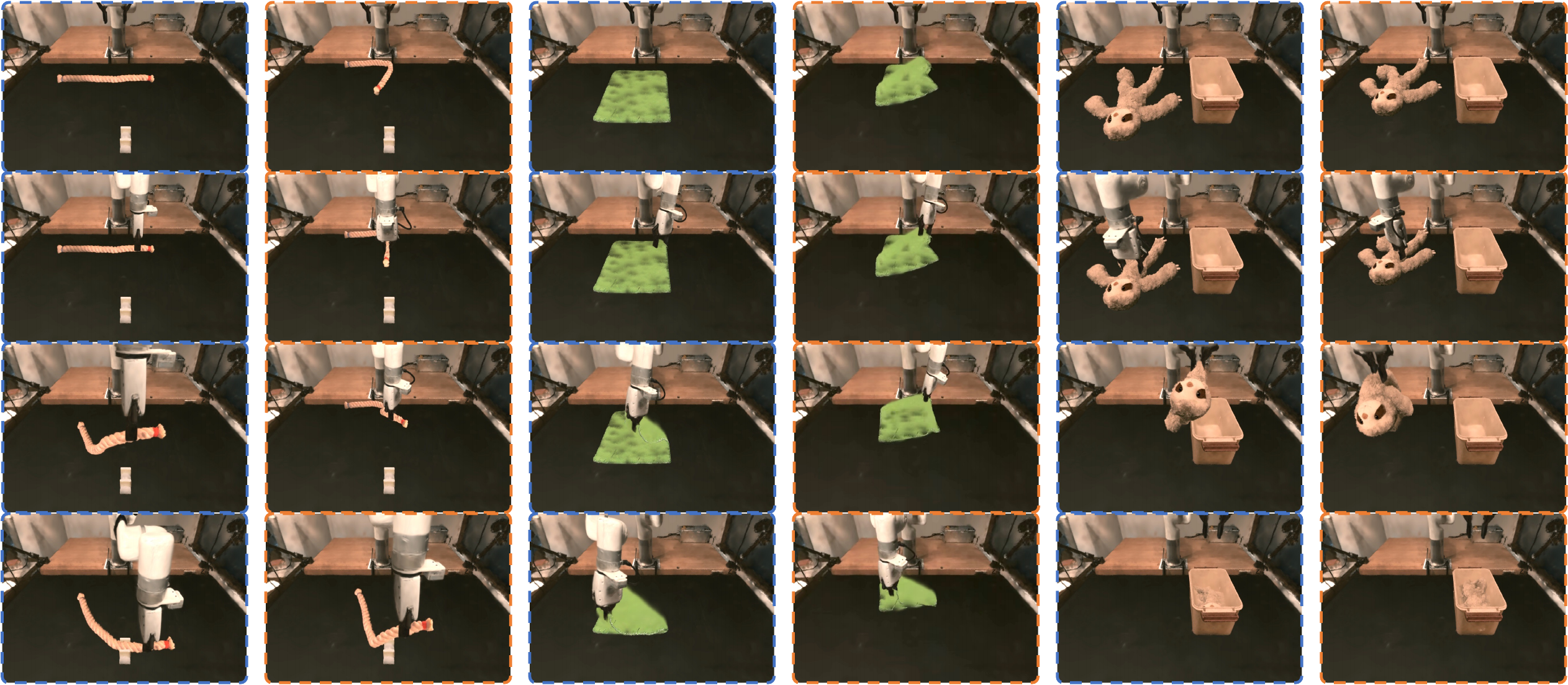}
    \caption{\textbf{Trajectory warping examples.} For each task, we show the source trajectory (blue) on the original object state and the warped trajectory (orange) on the augmented state. The deformation field adapts both the grasp pose and the manipulation path to the new geometry.}
    \label{fig:traj_warp_examples}
    \vspace{-4mm}
\end{figure}

\subsection{Policy Training}
\label{subsec:policy_learning}
To evaluate the effectiveness and efficiency of our augmentation approach, policies are trained via imitation learning and validated within a simulation environment. For three tasks---rope routing, toy packing, and cloth folding---we collect one teleoperation demonstration per task. Using the state augmentation method described in Sec.~\ref{subsec:state_augmentation}, we synthesize more than $1200$ distinct object states for each task to facilitate trajectory synthesis and evaluation scenarios. 
Subsequently, we employ the \textit{Deformation Field Warping} method and the local rigid-transfer ablation for trajectory augmentation.

The augmented trajectories are executed in simulation to verify their success, with task-specific success criteria detailed in Appendix~\ref{app:task_details}. We record third-person and wrist-mounted RGB images along with corresponding actions during execution. The successful episodes are split into training and held-out test sets (details in Sec.~\ref{sec:imple}).

Following the protocol of Real2Sim-Eval~\cite{zhang2025real2sim}, we train four policy architectures: ACT~\cite{zhao23act}, Diffusion Policy~\cite{chi2023diffusion}, SmolVLA~\cite{shukor2025smolvla}, and $\pi_0$~\cite{24pi0} (fine-tuned via LoRA). The trained policies are evaluated on held-out object states unseen during training, including configurations where the warping method failed to generate successful trajectories.

\section{Experiments}
\label{sec:exp}

\subsection{Implementation Details}
\label{sec:imple}

\begin{figure}[tbp]
    \centering
    \includegraphics[width=\textwidth]{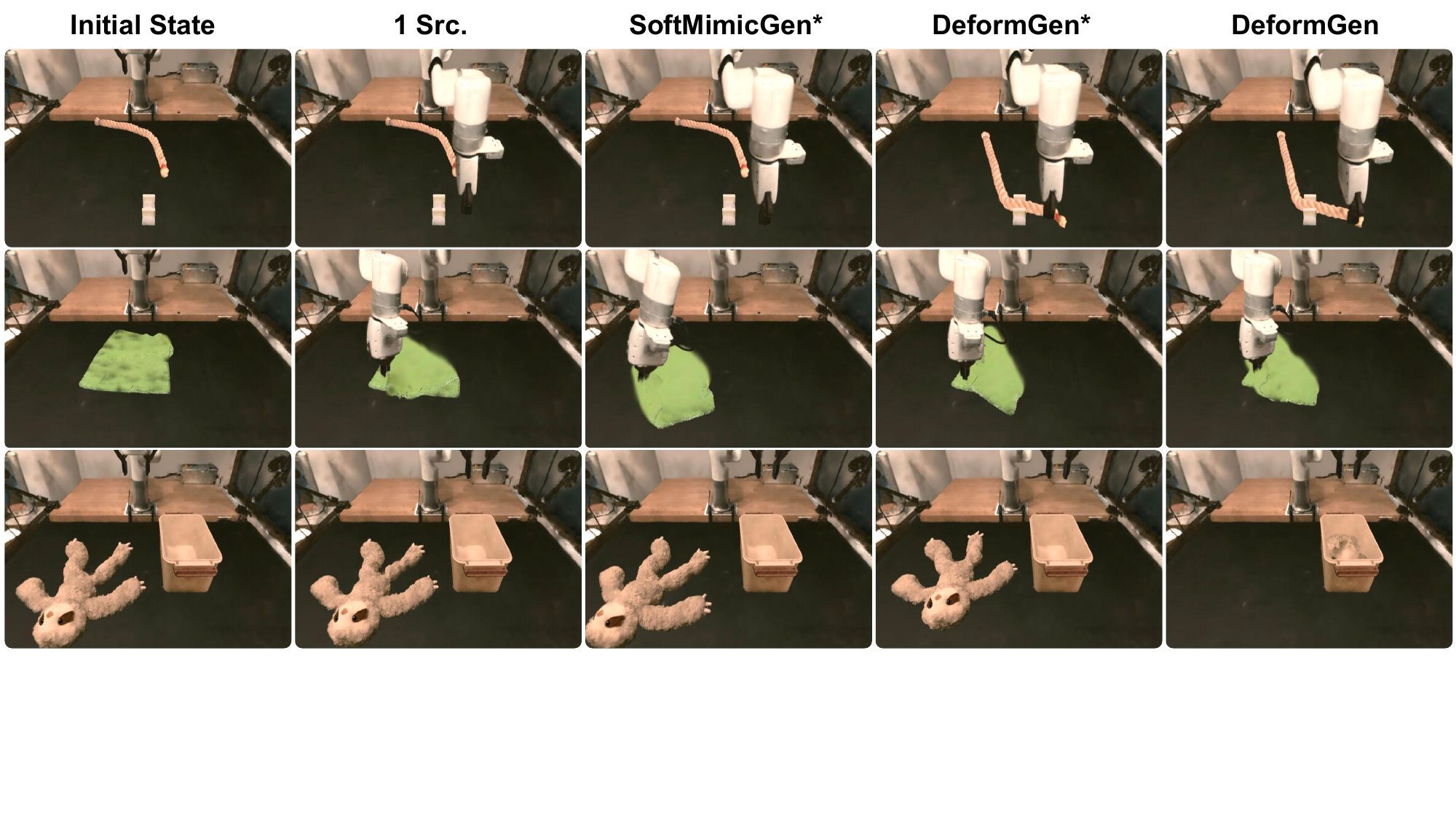}
    \vspace{-23mm}
    \caption{\textbf{Qualitative comparison of policy execution across methods.} Each row shows one task (rope routing, cloth folding, toy packing). Columns show the initial object state and the final rollout frame for policies trained under each regime. In these examples, \method{} consistently completes the task across diverse deformable configurations: threading the rope through the clip, folding the cloth into a triangle, and placing the toy into the container.}
    \label{fig:qualitative_comparison}
\end{figure}

All experiments are conducted in Real2Sim-Eval~\cite{zhang2025real2sim} with PhysTwin~\cite{jiang2025phystwin} for soft-body dynamics and rendering. 
The robot is an xArm7 with two RGB cameras (third-person and wrist, $848\times480$, 30\,Hz).
We evaluate on three tasks: \textbf{rope routing} (thread a rope through a clip), \textbf{toy packing} (place a stuffed toy into a container), and \textbf{cloth folding} (fold cloth into a triangle). Success criteria, augmentation parameters, and training hyperparameters are detailed in the Appendix.

For state augmentation, the gripper executes randomized Cartesian perturbations while in contact with the object (180 steps for rope/toy, 260 for cloth), followed by stabilization. 
For each task, we generate augmented states and attempt trajectory synthesis to obtain 1,000 successful trajectories for training and 200 successful states for testing. 

\paragraph{Compared methods.}
To disentangle the contributions of state augmentation and trajectory synthesis, we compare four training regimes:
\begin{itemize}[leftmargin=1.5em,topsep=2pt,itemsep=2pt]
\item \textbf{1\,Src.}: a single source demonstration without any augmentation.

\item \textbf{SoftMimicGen* (SMG*)}: SoftMimicGen~\cite{moghani2026softmimicgen} shares a similar philosophy to ours in trajectory synthesis---adapting demonstrations to deformed object geometry---but its state augmentation remains rigid: its state distribution is ``typically one with a larger set of possible placements for objects in the scene''~\cite{moghani2026softmimicgen}, i.e., SE(3) perturbations of object pose. Since SoftMimicGen is not open source, we reimplement its core design following the descriptions in the original paper. 

\item \textbf{DeformGen* (DG*)}: topological state augmentation paired with local rigid trajectory transfer. This ablation uses the same distribution of augmented states as DG but replaces deformation-field warping with a rigid transform estimated from the $K$ nearest material points around the grasp point and applied to the entire trajectory, isolating the effect of trajectory synthesis.

\item \textbf{DeformGen (DG)}: our full method, which pairs dynamics-based topological state augmentation with deformation-field warping. Compared with DG*, DG adapts the trajectory through a continuous deformation field rather than a single local rigid transform, enabling better alignment with the deformed object throughout manipulation.
\end{itemize}
The comparison between SMG* and DG reveals the effect of \emph{state augmentation} contrasting rigid with topological diversity, while the comparison between DG* and DG reveals the effect of \emph{trajectory synthesis} contrasting local rigid transfer with deformation-field warping.

\subsection{Experiment Results}
\label{sec:main_exp}

\definecolor{linecolor2}{RGB}{230, 234, 217}

Results are in Table~\ref{tab:main_results}. \method{} achieves the highest average success rate across three out of four policy architectures. The result reveals two insights:
\begin{table}[!htbp]
    \centering
    \scriptsize
    \caption{\textbf{Policy evaluation success rate} (\%) on deformable-object
    manipulation tasks. \textbf{1\,Src.}: 1 source demo. \textbf{SMG*}: rigid
    state aug.\ + deformation-field warping. \textbf{DG*}: topological state
  aug.\
    + local rigid transfer. \textbf{DG}: full \method.}
    \vspace{-3mm}
    \label{tab:main_results}
    \setlength{\tabcolsep}{2pt}
    \resizebox{\textwidth}{!}{%
    \begin{tabular}{lcccccccccccccccc}
    \toprule
    & \multicolumn{4}{c}{ACT~\cite{zhao23act}}
    & \multicolumn{4}{c}{DP~\cite{chi2023diffusion}}
    & \multicolumn{4}{c}{SmolVLA~\cite{shukor2025smolvla}}
    & \multicolumn{4}{c}{$\pi_{0}$~\cite{24pi0}} \\
    \cmidrule(lr){2-5}\cmidrule(lr){6-9}\cmidrule(lr){10-13}\cmidrule(lr){14-
  17}
    & 1 Src. & SMG* & DG* & \textbf{DG}
    & 1 Src. & SMG* & DG* & \textbf{DG}
    & 1 Src. & SMG* & DG* & \textbf{DG}
    & 1 Src. & SMG* & DG* & \textbf{DG} \\
    \midrule
    Rope
    & 0.00 & 68.00 & 90.00 & \cellcolor{linecolor2}\textbf{90.50}
    & 0.00 & \textbf{64.00} & \textbf{64.00} & \cellcolor{linecolor2}57.50
    & 0.00 & 62.50 & 88.00 & \cellcolor{linecolor2}\textbf{92.00}
    & 0.00 & 56.00 & 98.50 & \cellcolor{linecolor2}\textbf{99.00} \\
    Toy
    & 0.00 & 73.00 & 49.00 & \cellcolor{linecolor2}\textbf{75.50}
    & 0.00 & 49.50 & \textbf{56.50} & \cellcolor{linecolor2}54.00
    & 0.00 & 42.00 & 49.50 & \cellcolor{linecolor2}\textbf{53.50}
    & 0.00 & 10.00 & 32.50 & \cellcolor{linecolor2}\textbf{58.00} \\
    Cloth
    & 4.00 & 3.50 & 1.50 & \cellcolor{linecolor2}\textbf{11.00}
    & \textbf{7.00} & 0.50 & 2.50 & \cellcolor{linecolor2}0.50
    & 7.50 & 16.50 & \textbf{27.50} & \cellcolor{linecolor2}24.00
    & 7.00 & 17.50 & \textbf{24.00} & \cellcolor{linecolor2}13.00 \\
    \midrule
    Average
    & 1.33 & 48.17 & 46.83 & \cellcolor{linecolor2}\textbf{59.00}
    & 2.33 & 38.00 & \textbf{41.00} & \cellcolor{linecolor2}37.33
    & 2.50 & 40.33 & 55.00 & \cellcolor{linecolor2}\textbf{56.50}
    & 2.33 & 27.83 & 51.67 & \cellcolor{linecolor2}\textbf{56.67} \\
    \bottomrule
    \end{tabular}
    }
  \end{table}
  
\textbf{(1) Topological state diversity contributes to generalization.} Comparing SMG* with DG highlights the effect of state augmentation: both methods use deformation-field warping for trajectory transfer, but SMG* relies on rigid state perturbations whereas DG uses dynamics-based topological state augmentation. DG achieves higher average success rates in most architectures, suggesting that broader coverage of deformable-object configurations is important for policy generalization.

\textbf{(2) Deformation-field warping provides complementary gains.} Since DG* and DG use the same topologically augmented states, their comparison isolates trajectory transfer. DG often improves over DG*, suggesting that deformation-aware warping can further benefit policy learning.

Together, these findings suggest that both broader deformable-state coverage and deformation-aware trajectory transfer contribute to policy generalization, with their relative effects varying across architectures and tasks.
Figure~\ref{fig:qualitative_comparison} qualitatively illustrates this trend: policies trained with \method{} successfully complete tasks while comparison methods often exhibit grasp misalignment or incomplete manipulation.

\begin{figure}[!htbp]
    \centering
    \includegraphics[width=0.95\textwidth]{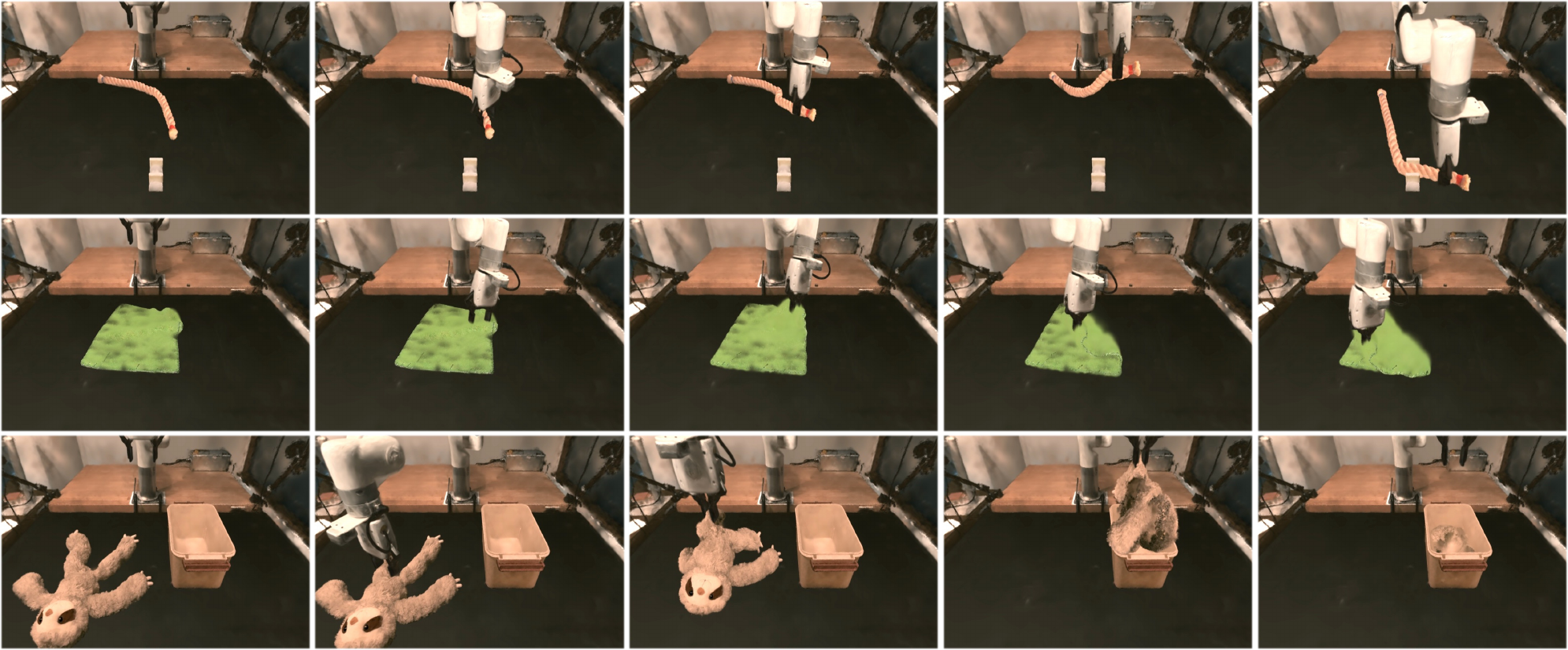}
    \caption{\textbf{Policy execution rollouts on unseen test states.} Each row shows key frames from a successful episode. The policies are trained on \method{}-augmented data and evaluated on held-out object configurations not seen during training.}
    \label{fig:rollout_examples}
    \vspace{-4mm}
\end{figure}

\subsection{State Coverage Analysis}
\label{sec:state_analysis}

\begin{figure}[!htbp]
    \centering
    \includegraphics[width=1.0\textwidth]{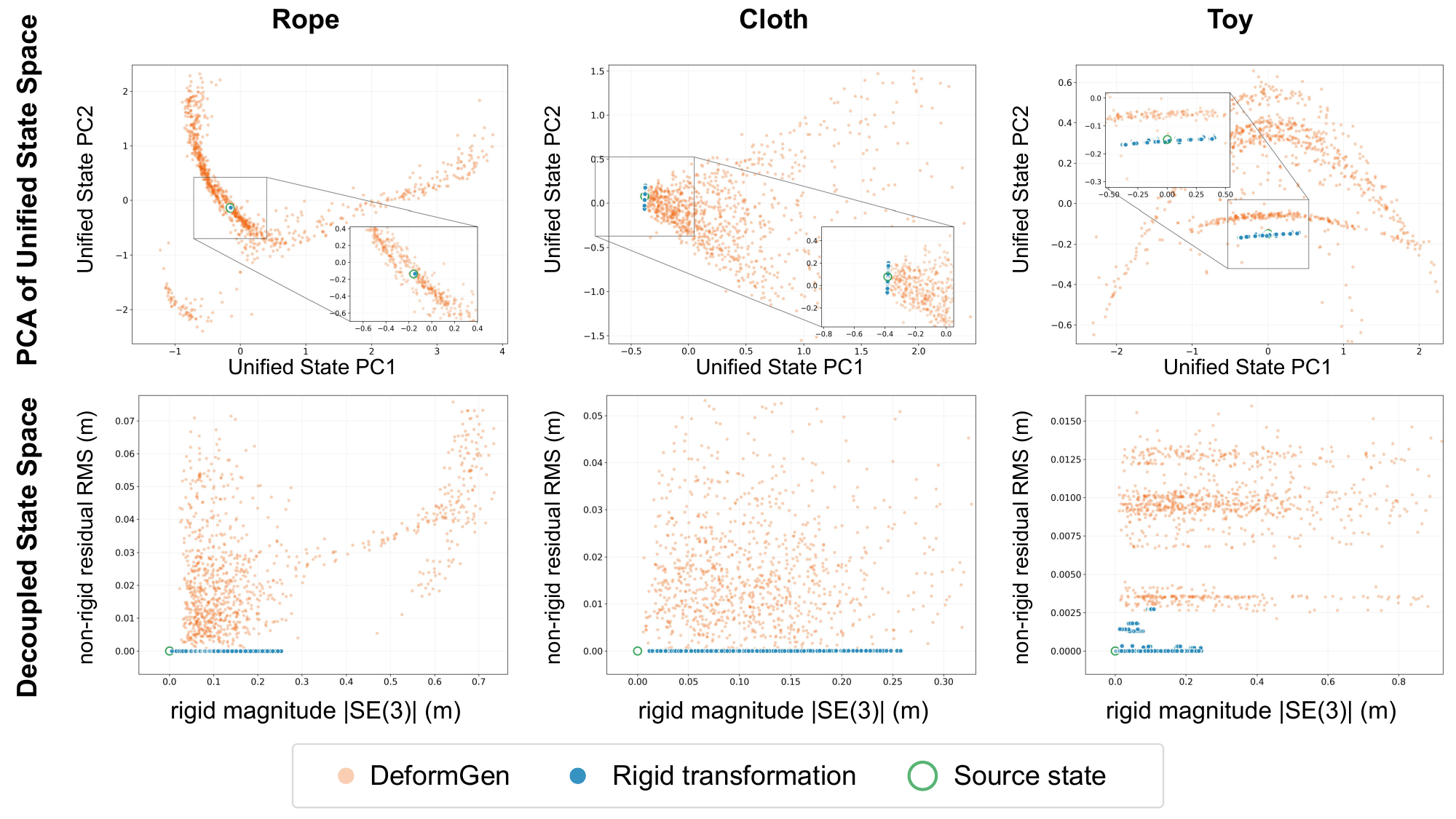}
    \vspace{-6mm}
    \caption{
\textbf{State-space analysis across three tasks.} Each state is decomposed relative to the source (\textcolor{green}{green circle}) into a rigid $\mathrm{SE}(3)$ component and a non-rigid residual. \textbf{Top:} PCA of the unified state vector. Rigid samples (\textcolor{blue}{blue}) cluster near the source; \method (\textcolor{orange}{orange}) spreads broadly. \textbf{Bottom:} Rigid magnitude ($x$) vs.\ non-rigid residual RMS ($y$). Rigid samples have near-zero residual; \method{}  shows large residuals confirming genuine shape deformations. The toy case has non-zero residuals due to deformation from object interactions during stabilization.}
    \label{fig:state_analysis}
\vspace{-5mm}
\end{figure}

We decompose each augmented state into a rigid SE(3) component and a non-rigid residual via Procrustes alignment (Figure~\ref{fig:state_analysis}). Rigid augmentation clusters near the source with negligible residual, while \method{} spreads broadly with large residuals---confirming that the performance gains in Table~\ref{tab:main_results} stem from genuine topological diversity rather than merely more data at similar configurations.

Figure~\ref{fig:state_aug_examples} shows representative augmented object states generated by DeformGen for each task. Starting from a single source configuration, our dynamics-based augmentation produces a diverse set of topologically distinct states---including different rope curvatures, varied stuffed toy orientations and compressions, and diverse cloth folds and drapes---all of which are physically plausible under the simulator's dynamics model.

\subsection{Ablation Studies}
\label{sec:ab_exp}

\question{\bf Can \method{} generalize to rigid-only scenarios?}
Table~\ref{tab:rigid_test_results} evaluates whether topological augmentation hurts performance when the test states only involve rigid transformations. SMG* achieves the highest average success rate for ACT, DP, and SmolVLA, which is expected because its rigid-state training distribution closely matches the rigid-only test set. In contrast, DG* and DG are trained on broader topological variations. Nevertheless, DG remains competitive and even performs best for $\pi_0$, suggesting that training on topologically diverse data does not substantially compromise performance on simpler rigid scenarios.

\begin{table}[!htbp]
  \centering
  \small
  \caption{Success rate (\%) on \textbf{rigid-only} test states.
  }
  \label{tab:rigid_test_results}
  \resizebox{\columnwidth}{!}{%
  \setlength{\tabcolsep}{6pt}
  \begin{tabular}{lcccccccccccc}
  \toprule
  Task
  & \multicolumn{3}{c}{ACT}
  & \multicolumn{3}{c}{DP}
  & \multicolumn{3}{c}{SmolVLA}
  & \multicolumn{3}{c}{$\pi_0$} \\
  \cmidrule(lr){2-4}\cmidrule(lr){5-7}\cmidrule(lr){8-10}\cmidrule(lr){11-13}
  & SMG* & DG* & DG
  & SMG* & DG* & DG
  & SMG* & DG* & DG
  & SMG* & DG* & DG \\
  \midrule
  Rope
  & \textbf{94.00} & 78.50 & 77.50
  & \textbf{92.50} & 51.50 & 46.00
  & \textbf{91.00} & 69.00 & 62.00
  & 92.00 & \textbf{95.50} & 86.50 \\
  Toy
  & \textbf{90.00} & 76.00 & 83.50
  & 77.50 & \textbf{84.50} & 69.00
  & 45.50 & 47.00 & \textbf{57.00}
  & 12.50 & 29.00 & \textbf{39.50} \\
  Cloth
  & 6.50 & 10.00 & \textbf{13.00}
  & 2.00 & \textbf{7.50} & 7.00
  & 27.00 & 20.50 & \textbf{30.50}
  & 8.50 & 20.50 & \textbf{34.50} \\
  \midrule
  Average
  & \textbf{63.50} & 54.83 & 58.00
  & \textbf{57.33} & 47.83 & 40.67
  & \textbf{54.50} & 45.50 & 49.83
  & 37.67 & 48.33 & \textbf{53.50} \\
  \bottomrule
  \end{tabular}%
  }
  \end{table}

\question{\bf Impact of synthetic data quantity.} Table~\ref{tab:datascale_results} shows that average performance improves with scale. The average success rates increase monotonically from $N{=}100$ to $N{=}750$ for both ACT (19.50\% $\to$ 61.50\%) and SmolVLA (36.83\% $\to$ 63.17\%). This suggests that dynamics-based augmentation can benefit from increased data scale, with larger synthetic datasets providing average gains.

\begin{table}[!htbp]
\centering
\small
\caption{Success rate (\%) under different synthetic data quantity.}
\label{tab:datascale_results}
\setlength{\tabcolsep}{3pt}
\begin{tabular*}{\columnwidth}{@{\extracolsep{\fill}}lcccccccc@{}}
\toprule
& \multicolumn{4}{c}{ACT}
& \multicolumn{4}{c}{SmolVLA} \\
\cmidrule(lr){2-5}\cmidrule(lr){6-9}
& N = 100 & N = 250 & N = 500 & N = 750
& N = 100 & N = 250 & N = 500 & N = 750 \\
\midrule
Rope
& 36.50 & 57.50 & 75.50 & \textbf{88.50}
& 65.00  & 80.50 & 84.00 & \textbf{91.00}\\
Toy
& 14.50 &  \textbf{82.50} & 79.00 & 74.00
& 20.00  & 28.00 & 59.50 & \textbf{66.00}\\
Cloth
& 7.50 & 14.50 & 21.50 & \textbf{22.00}
& 25.50  & 5.50 & 24.00 & \textbf{32.50}\\
\midrule
Average
& 19.50 & 51.50 & 58.67 & \textbf{61.50}
& 36.83 & 38.00 & 55.83 & \textbf{63.17} \\
\bottomrule
\end{tabular*}
\end{table}

\question{\bf Can the policy generalize to synthesis failure cases?}
This ablation tests whether the policy merely memorizes the augmented trajectories or learns transferable manipulation strategies. We evaluate on \textit{hard samples}---states where trajectory synthesis itself failed to produce a valid demonstration, meaning the policy has never seen a successful trajectory for these configurations. Since rope achieves nearly 100\% synthesis success, this study is conducted on toy and cloth only.
As shown in Table~\ref{tab:hard_case_results}, policies trained with augmentation achieve non-trivial success on these out-of-distribution states, though performance varies across tasks and architectures. Augmentation helps policies learn generalizable manipulation strategies rather than memorizing individual demonstrations, enabling some degree of extrapolation to unseen configurations.

\begin{table}[h]
\centering
\small
\caption{Success rate (\%) on \textbf{hard samples} where synthesis failed.
Rope excluded (nearly 100\% synthesis success).
\textbf{SMG*}: rigid state aug. \textbf{DG*}: topological state + rigid
transfer.
\textbf{DG}: full \method.}
\label{tab:hard_case_results}
\resizebox{\columnwidth}{!}{%
\begin{tabular}{lcccccccccccc}
\toprule
Task
& \multicolumn{3}{c}{ACT}
& \multicolumn{3}{c}{DP}
& \multicolumn{3}{c}{SmolVLA}
& \multicolumn{3}{c}{$\pi_0$} \\
\cmidrule(lr){2-4}\cmidrule(lr){5-7}\cmidrule(lr){8-10}\cmidrule(lr){11-13}
& SMG* & DG* & DG
& SMG* & DG* & DG
& SMG* & DG* & DG
& SMG* & DG* & DG \\
\midrule
Toy
& 45.50 & 37.50 & \textbf{55.50}
& 37.50 & \textbf{55.00} & 47.00
& 15.00 & 15.00 & \textbf{18.50}
& 7.50 & 18.50 & \textbf{36.50} \\
Cloth
& \textbf{11.00} & 6.00 & 5.50
& \textbf{6.00} & 5.00 & 5.00
& \textbf{17.50} & 16.50 & 10.00
& 11.50 & \textbf{19.00} & 4.00 \\
\midrule
Average
& 28.25 & 21.75 & \textbf{30.50}
& 21.75 & \textbf{30.00} & 26.00
& \textbf{16.25} & 15.75 & 14.25
& 9.50 & 18.75 & \textbf{20.25} \\
\bottomrule
\end{tabular}%
}
\vspace{-3mm}
\end{table}

\subsection{Failure Analysis}
\label{sec:failure_analysis}

\begin{figure}[tbp]
    \centering
    \includegraphics[width=0.9\textwidth]{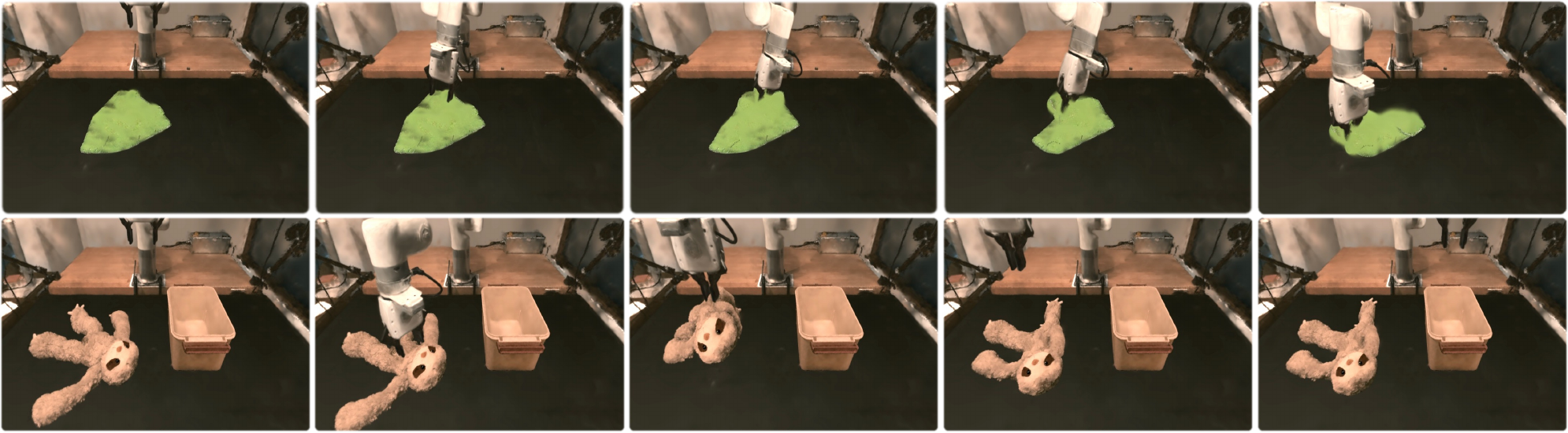}
    \caption{\textbf{Representative failure cases.} Common failure modes include grasp misalignment on extreme deformations (Top), and premature release due to contact instability (Bottom).}
    \label{fig:failure_cases}
    \vspace{-4mm}
\end{figure}

Figure~\ref{fig:failure_cases} shows representative failure cases. Common failure modes include imprecise grasp on highly deformed configurations where the visual appearance deviates significantly from training data, and premature release due to unstable contact under large deformations.

\section{Conclusion}
In this work, we proposed \method, a dynamics-based augmentation framework that expands the valid state distribution through localized physical disturbances, forward simulation, and stabilization, and transfers source manipulation trajectories via deformation-field warping. In this way, DeformGen augments both deformable object states and their associated manipulation behaviors. Experiments on high-fidelity deformable manipulation benchmarks showed that DeformGen generally improves policy learning over both training on the original demonstrations alone and rigid-style augmentation baselines. More broadly, our results suggest that effective augmentation for deformable manipulation requires dynamics-consistent state synthesis and deformation-aware trajectory transfer.

\clearpage
\setlength{\bibsep}{5pt}
\bibliography{main}
\bibliographystyle{unsrtnat}

\newpage
\section*{Appendix}
\appendix
\section{State Augmentation Details}
\label{app:state_aug_details}

\subsection{Formal Assumption}
\label{app:formal_assumption}

Our approach relies on the premise that a well-calibrated physics simulator approximately preserves physical plausibility when evolving from a valid initial state. We formalize this as a working assumption on the simulator $\Phi_{\mathrm{sim}}(\mathbf{s}, \mathbf{f}, \Delta t)$, which evolves state $\mathbf{s}$ under external forces $\mathbf{f}$ over time interval $\Delta t$:

\begin{assumption}[Approximate conditional closure of $\mathcal{S}_{\mathrm{real}}$]
\label{ass:forward-invariance}
A sufficiently accurate physics simulator approximately preserves physical plausibility when starting from a valid state, whereas it cannot reliably restore plausibility from an out-of-distribution configuration:
\begin{align}
  \mathbf{s} \in \mathcal{S}_{\mathrm{real}}
  &\;\implies\;
  \Phi_{\mathrm{sim}}(\mathbf{s},\, \mathbf{f},\, \Delta t)
    \approx_{\mathcal{S}_{\mathrm{real}}},
  \quad
  \text{for reasonable } \mathbf{f} \text{ and } \Delta t;
  \tag{A1} \label{eq:A1} \\[4pt]
  \mathbf{s} \notin \mathcal{S}_{\mathrm{real}}
  &\;\not\!\!\!\implies\;
  \Phi_{\mathrm{sim}}(\mathbf{s},\, \mathbf{f},\, \Delta t)
    \in \mathcal{S}_{\mathrm{real}}.
  \tag{A2} \label{eq:A2}
\end{align}
\end{assumption}

\noindent
We note that this assumption is an idealization: real simulators introduce numerical integration errors and may not perfectly model all material properties, so the generated states are plausible \emph{with respect to the simulator's dynamics model} rather than guaranteed to match real-world physics exactly. The use of high-fidelity simulators (PhysTwin~\cite{jiang2025phystwin}) narrows this gap in practice.

\subsection{Detailed Analysis of Existing Strategies}
\label{app:augmentation_analysis}

We provide a detailed analysis of three representative augmentation strategies and their limitations when applied to deformable objects.

\paragraph{(i) Global rigid transformation.}
Applying a uniform $\mathbf{T} \in SE(3)$ to all particles preserves all inter-particle relations, so the augmented state remains in $\mathcal{S}_{\mathrm{real}}$. However, the reachable set is confined to a 6-dimensional subspace spanned by rigid pose variations of~$\mathbf{s}_0$, which cannot capture any shape or topological variation of deformable objects. This is confirmed empirically in Fig.~\ref{fig:state_analysis}: rigid augmentation produces near-zero non-rigid residuals across all three tasks.

\paragraph{(ii) Per-particle independent perturbation.}
Adding independent noise $\boldsymbol{\epsilon}_i \sim \mathcal{P}(\sigma)$ to each particle can in principle reach any $\mathbf{s} \in \mathcal{S}$, but faces a practical coverage--plausibility trade-off. Large~$\sigma$ produces disordered point clouds that break inter-particle connectivity and introduce topology artifacts (e.g., self-intersections, disconnected segments), pushing the state far outside~$\mathcal{S}_{\mathrm{real}}$ in ways that subsequent stabilization steps typically cannot recover. Small~$\sigma$ preserves local topology but induces only surface wrinkles~\cite{tian2025interndata}, confining the resulting states to a local neighborhood of~$\mathbf{s}_0$ with insufficient diversity for policy learning. Crucially, even when a stabilization step is applied, there is no mechanism to verify whether the result has returned to~$\mathcal{S}_{\mathrm{real}}$, making this approach unreliable in practice.

\paragraph{(iii) Kinematic topological transformation.}
Applying a continuous deformation field $\boldsymbol{\phi}: \mathbb{R}^3 \to \mathbb{R}^3$ improves upon~(ii) by preserving topological coherence---the connectivity structure is maintained by construction. However, $\boldsymbol{\phi}$ is constructed without reference to the object's material model, so the deformed state may violate internal dynamic constraints (e.g., producing rest-shape configurations with unrealistic internal stress or interpenetration with the environment). These configurations are structurally coherent but dynamically inadmissible, and stabilization cannot reliably project them back onto~$\mathcal{S}_{\mathrm{real}}$ because the simulator's corrective dynamics may converge to a different basin or fail to converge at all.

\subsection{Advantages of Dynamics-Based Augmentation over (ii) and (iii)}
\label{app:dynamics_advantages}

Beyond broader coverage relative to rigid transformations, our dynamics-based approach offers two practical advantages over per-particle perturbation and kinematic deformation:

\begin{itemize}[leftmargin=*,itemsep=4pt]
    \item \textbf{No plausibility--diversity trade-off.} Both strategies~(ii) and~(iii) face a fundamental tension: increasing perturbation magnitude increases diversity but also the likelihood of producing implausible configurations. Our method sidesteps this trade-off because diversity is achieved through the simulator's own dynamics---larger or longer-duration forces naturally produce more diverse states, while the simulation's internal constraints (collision handling, material constitutive laws, boundary conditions) continuously enforce plausibility throughout the trajectory. There is no separate perturbation-then-repair pipeline that could fail.
    
    \item \textbf{Implicit enforcement of coupled constraints.} Deformable objects are subject to multiple interacting constraints simultaneously: material elasticity, self-collision avoidance, environmental contact, and gravitational settling. Strategies~(ii) and~(iii) perturb geometry without awareness of these coupled constraints, and a subsequent stabilization step can at best enforce them approximately and sequentially. In contrast, forward simulation enforces all constraints jointly at each time step through the simulator's integrated solver, producing states where internal stresses, contact forces, and boundary conditions are mutually consistent. This is particularly important for objects with complex rest-state interactions (e.g., a rope draped over a fixture, or cloth resting on a surface with folds), where violating one constraint easily cascades into violations of others.
\end{itemize}

\subsection{Reachable Set Discussion}
\label{app:reachable_set}

The set of states reachable via dynamics simulation from~$\mathbf{s}_0$ is:
\begin{equation}
    \mathcal{R}(\mathbf{s}_0) \;=\; \left\{
      \Phi_{\mathrm{sim}}(\mathbf{s}_0,\, \mathbf{f},\, \Delta t)
      \;\middle|\;
      \mathbf{f} \in \mathbb{R}^{3N},\;
      \Delta t > 0
    \right\}.
  \label{eq:reachable}
\end{equation}
In principle, $\mathcal{R}(\mathbf{s}_0)$ could be large, since any physically plausible configuration is connected to~$\mathbf{s}_0$ through some physical process. However, we do not claim that our randomized sampling of~$(\mathbf{f}, \Delta t)$ achieves full coverage of~$\mathcal{S}_{\mathrm{real}}$ or matches the true distribution of real-world configurations. We treat dynamics-based augmentation as a practical sampling heuristic that explores a substantially broader and more physically grounded region of the state space than rigid transformations. This is verified empirically in Sec.~\ref{sec:state_analysis}.

\section{Trajectory Augmentation Details}
\label{app:trajectory_details}

\subsection{Decay Function}
\label{app:decay_function}

The decay function $\alpha_t = \mathrm{decay}(t)$ in the position and orientation warping controls how strongly the deformation field influences the trajectory over time. We support three configurations:
\begin{itemize}[leftmargin=*,itemsep=2pt]
    \item \textit{None}: $\alpha_t = 1$ for all $t$. The deformation field is applied uniformly throughout the trajectory.
    \item \textit{Linear}: $\alpha_t = \max(0,\; 1 - t/T)$, where $T$ is the total trajectory length. The influence decreases linearly to zero.
    \item \textit{Exponential}: $\alpha_t = e^{-\lambda t}$, where $\lambda > 0$ controls the decay rate. The influence decreases exponentially.
\end{itemize}
The decay allows the trajectory to closely follow local deformations near the grasp phase while gradually reverting to the original trajectory towards the end of the manipulation phase. The choice of decay function is task-dependent and specified in Appendix~\ref{app:task_hyperparams}.

\subsection{KNN Scope for Grasp vs.\ Manipulation Phases}
\label{app:knn_scope}

In practice, we observe that the grasp pose correlates more strongly with object points in the vicinity of the grasp point. Therefore, we employ a small $K$ (e.g., $K = 5$--$10$) for warping the grasp pose, so that only nearby particle displacements influence the grasp alignment.

Conversely, the manipulation phase depends on the overall object state; the end-effector must compensate not only for local geometry changes but also for global shape shifts. We therefore set $K$ equal to the total number of object points $N$ for the manipulation trajectory, effectively using a globally weighted deformation field.

\subsection{Orientation Constraints}
\label{app:orientation_constraints}

Given the tabletop manipulation scenario in our experiments, significant orientation changes occur primarily around the Z-axis (perpendicular to the table surface). We therefore constrain the orientation warping to the Z-axis component: the original rotation matrix $R_t$ and the induced rotation $R_t'$ are first projected onto their Z-axis rotational components before SLERP interpolation is applied. This prevents spurious tilting or flipping of the end-effector that could arise from noisy Jacobian estimates in the other axes.

\section{Implementation Details}
\label{app:impl}

\subsection{Simulation and Robot Setup}
\label{app:simulation_details}

All experiments are conducted in Real2Sim-Eval~\cite{zhang2025real2sim}, which provides physically accurate soft-body dynamics and photorealistic rendering via PhysTwin~\cite{jiang2025phystwin}. The robot is an xArm7 manipulator equipped with two RGB cameras: a fixed third-person camera and a wrist-mounted camera, both at $848\times480$ resolution and 30\,Hz frame rate. The policy outputs 8-dimensional actions consisting of end-effector position $(x, y, z)$, quaternion orientation $(q_w, q_x, q_y, q_z)$, and gripper opening at 30\,Hz control frequency. Internally, the simulation converts the policy output to a 13D command (xyz $+$ $3\times3$ rotation matrix $+$ gripper) before execution.

\subsection{Task Descriptions and Success Criteria}
\label{app:task_details}

\paragraph{Rope routing.}
The robot must thread a deformable rope through a clip. Success is evaluated over the final 100 frames of each episode. The episode is considered successful if at least 30 frames satisfy the condition that the rope forms sufficient intersections (at least 100 spring-segment crossings) with both the upper and lower planes of the clip, indicating that the rope has been threaded through.

\paragraph{Toy packing.}
The robot must place a stuffed toy into a container. Success is evaluated at the final frame. We construct a minimum oriented bounding box (OBB) from the initial reference mesh and scale it by a factor of $1.05$. The episode succeeds if at least 3{,}050 object points fall within this scaled OBB.

\paragraph{Cloth folding.}
The robot must fold a cloth into a triangular shape. Success is evaluated at the final frame. The point cloud is projected onto the table plane to form a binary mask. We extract the largest connected component, fit a minimum bounding triangle, and verify three conditions: (i)~the contour has 3--4 approximate vertices, (ii)~the IoU between the mask and the fitted triangle $\geq 0.72$, and (iii)~the mask coverage of the triangle $\geq 0.80$.

\subsection{State Augmentation Parameters}
\label{app:state_aug_params}

Since the simulation environment does not expose a direct external-force API, we implement localized physical disturbances by commanding the gripper to execute randomized Cartesian perturbations while in contact with the object, transmitting forces through contact dynamics. Each augmentation episode consists of a sequence of random steps; each step applies either a planar translation (sampled from discrete $\pm x$, $\pm y$ directions) or a $z$-axis rotation (with probability $p_{\mathrm{rot}}$). Task-specific configurations are as follows:

\begin{itemize}[leftmargin=1.5em,topsep=2pt,itemsep=2pt]
\item \textbf{Rope / Toy}: 180 random steps, translation magnitudes sampled from $\{0.012, 0.006, 0.003\}$\,m, rotation steps of $\pm6^{\circ}$, rotation probability $p_{\mathrm{rot}} = 0.45$.
\item \textbf{Cloth}: 260 random steps, translation magnitudes sampled from $\{0.018, 0.009, 0.0045\}$\,m, rotation steps of $\pm8^{\circ}$, rotation probability $p_{\mathrm{rot}} = 0.55$.
\end{itemize}

After perturbation, the object is stabilized for 30--40 simulation steps to reach quasi-static equilibrium.

\subsection{Data Splits}
\label{app:data_splits}

For each task, we generate augmented states via dynamics-based topological transformation and attempt trajectory synthesis until obtaining sufficient successful demonstrations. The generation statistics are:
\begin{itemize}[leftmargin=1.5em,topsep=2pt,itemsep=2pt]
\item \textbf{Rope}: 1,294 successful trajectories out of 1,300 generated states. Trajectory synthesis success rate: 99.5\%. Split into 1,000 training / 200 test. Remaining successful trajectories are unused.
\item \textbf{Toy}: 1,327 successful trajectories out of 2,200 generated states. Trajectory synthesis success rate: 60.3\%. Split into 1,000 training / 200 test, with 200 failed states sampled from 873 available. Remaining successful trajectories are unused.
\item \textbf{Cloth}: 1,778 successful trajectories out of 4,500 generated states. Trajectory synthesis success rate: 39.5\%. Split into 1,000 training / 200 test, with 200 failed states sampled from 2,722 available. Remaining successful trajectories are unused.
\end{itemize}
Failed states from toy and cloth tasks serve as \textit{hard samples} for the generalization ablation in Sec.~\ref{sec:ab_exp}. Rope is excluded from this ablation due to insufficient failed samples.

\subsection{Trajectory Augmentation Hyperparameters}
\label{app:task_hyperparams}

Task-specific trajectory warping configurations:
\begin{itemize}[leftmargin=1.5em,topsep=2pt,itemsep=2pt]
\item \textbf{Rope}: Grasp KNN $K=5$, manipulation KNN $K=N$ (all points), decay function: linear.
\item \textbf{Toy}: Grasp KNN $K=5$, manipulation KNN $K=N$, decay function: none.
\item \textbf{Cloth}: Grasp KNN $K=10$, manipulation KNN $K=N$, decay function: exponential ($\lambda=0.02$).
\end{itemize}

\subsection{Policy Training Hyperparameters}
\label{app:policy_hyperparams}

All policies are trained on a single NVIDIA A100 GPU. Hyperparameters are tuned per algorithm:

\paragraph{ACT~\cite{zhao23act}.}
Learning rate: $1\times10^{-5}$. Batch size: 512. Training epochs: 10. 

\paragraph{Diffusion Policy (DP)~\cite{chi2023diffusion}.}
Learning rate: $1\times10^{-4}$. Batch size: 512. Training epochs: 10. Scheduler: cosine with 500-step warmup.

\paragraph{SmolVLA~\cite{shukor2025smolvla}.}
Learning rate: $1\times10^{-4}$. Batch size: 128. Training epochs: 10. Scheduler: warmup 1000.

\paragraph{$\pi_0$~\cite{24pi0}.}
Fine-tuned via LoRA using the OpenPI framework. Peak learning rate: $2.5\times10^{-5}$, decay learning rate: $2.5\times10^{-6}$. Batch size: 8. Training epochs: 10. Optimizer: AdamW ($\beta_1{=}0.9$, $\beta_2{=}0.95$, weight decay $1\times10^{-10}$). Scheduler: cosine decay.

\section{Limitations}
\label{app:limitations}

We acknowledge several limitations of the current work:

\begin{itemize}[leftmargin=1.5em,topsep=4pt,itemsep=4pt]

\item \textbf{Single-arm manipulation only.} All experiments are conducted with a single xArm7 manipulator. Extending DeformGen to bimanual or multi-robot settings---where coordination between arms introduces additional constraints on trajectory synthesis---remains future work.

\item \textbf{Limited task diversity.} We validate on three deformable manipulation tasks (rope, stuffed toy, cloth), which cover a range of material properties (1D, quasi-rigid 3D, 2D sheet). However, other important categories such as dough/clay shaping, surgical tissue manipulation, or cable routing in cluttered environments have not been evaluated. The generality of our dynamics-based augmentation to these domains remains to be demonstrated.

\item \textbf{Sim-to-real gap.} All experiments are conducted entirely in simulation. While Real2Sim-Eval and PhysTwin provide high-fidelity physics and rendering, transferring the augmented policies to real hardware may require additional domain adaptation or fine-tuning to handle discrepancies in material properties, contact dynamics, and visual appearance.

\item \textbf{Trajectory synthesis is not universally successful.} Guaranteeing successful trajectory transfer for arbitrary initial states is inherently difficult. Following the core philosophy of~\cite{schulman2016learning}, our warping assumes that geometric correspondence preserves task semantics, but this only holds approximately---complex contact dynamics, large topological changes, and kinematic constraints can cause warped trajectories to fail. This is reflected in our varying success rates. Encouragingly, policies trained on successful trajectories still generalize to some failure states (Sec.~\ref{sec:ab_exp}). Future work could mitigate this via iterative human-in-the-loop demonstration collection, closed-loop trajectory refinement, or multi-source warping that selects the most compatible demonstration for each target state.

\end{itemize}

\section{Broader Impacts}
\label{app:broader_impacts}

This work aims to reduce the cost of collecting manipulation demonstrations for deformable objects by providing an automated data augmentation pipeline. The positive societal impact includes enabling more accessible and scalable robot learning for tasks involving soft materials (e.g., household assistance, garment handling, food preparation), potentially benefiting applications in elder care and manufacturing.

As the method operates entirely in simulation for data synthesis and does not involve real-world data collection, human subjects, or generation of potentially harmful content, we do not foresee direct negative societal impacts beyond standard safety considerations for robotic manipulation systems. The trained policies are task-specific manipulation controllers without broader capabilities that could be misused.

\section{Licenses}
\label{app:licenses}

We list the licenses of all external assets used in this work:

\begin{itemize}[leftmargin=1.5em,topsep=2pt,itemsep=2pt]
\item \textbf{Real2Sim-Eval}~\cite{zhang2025real2sim}: MIT License
\item \textbf{PhysTwin}~\cite{jiang2025phystwin}: MIT License
\item \textbf{ACT}~\cite{zhao23act}: MIT License
\item \textbf{Diffusion Policy}~\cite{chi2023diffusion}: MIT License
\item \textbf{SmolVLA}~\cite{shukor2025smolvla}: Apache License 2.0
\item \textbf{$\pi_0$}~\cite{24pi0}: Apache License 2.0
\end{itemize}

Our use of these assets complies with their respective license terms.

\end{document}